\pdfoutput=1

\documentclass[11pt]{article}

\usepackage[]{EMNLP2023}

\usepackage{times}
\usepackage{latexsym}

\usepackage[T1]{fontenc}

\usepackage[utf8]{inputenc}

\usepackage{microtype}

\usepackage{inconsolata}

\usepackage{graphicx}
\usepackage{multirow}
\usepackage{subcaption}
\usepackage{booktabs}

\title{Reducing Sequence Length by Predicting Edit Spans with Large Language Models}

\author{Masahiro Kaneko$^{1,2}$ \quad
        Naoaki Okazaki$^{2}$ \\
        $^1$MBZUAI \\
        $^2$Tokyo Institute of Technology \\
        {\tt Masahiro.Kaneko@mbzuai.ac.ae} \quad
        {\tt okazaki@c.titech.ac.jp}}

\begin{document}
\maketitle
\begin{abstract}
Large Language Models (LLMs) have demonstrated remarkable performance in various tasks and gained significant attention.
LLMs are also used for local sequence transduction tasks, including grammatical error correction (GEC) and formality style transfer, where most tokens in a source text are kept unchanged.
However, the models that generate all target tokens in such tasks have a tendency to simply copy the input text as is, without making needed changes, because the difference between input and output texts is minimal in the training data.
This is also inefficient because the computational cost grows quadratically with the target sequence length with Transformer.
This paper proposes predicting edit spans for the source text for local sequence transduction tasks.
Representing an edit span with a position of the source text and corrected tokens, we can reduce the length of the target sequence and the computational cost for inference.
We apply instruction tuning for LLMs on the supervision data of edit spans.
Experiments show that the proposed method achieves comparable performance to the baseline in four tasks, paraphrasing, formality style transfer, GEC, and text simplification, despite reducing the length of the target text by as small as 21\%.
Furthermore, we report that the task-specific fine-tuning with the proposed method achieved state-of-the-art performance in the four tasks.
\end{abstract}

\section{Introduction}
\label{sec:intro}

Large Language Models (LLMs), including ChatGPT\footnote{\url{https://chat.openai.com/}} and Bard\footnote{\url{https://bard.google.com/}}, have exhibited exceptional performance across a range of natural language processing (NLP) tasks and amassed a significant user base~\cite{NEURIPS2020_1457c0d6,chowdhery2022palm,chatgpt}.
As performance gains are brought from the increases in model size~\cite{kaplan2020scaling,wei2022emergent,zhao2023survey}, LLMs are becoming larger and larger.
However, the computational cost of inference is a severe bottleneck of many practical applications, especially when the number of parameters in an LLM is massive~\cite{bender2021dangers,kraus2023enhancing}.

Meanwhile, LLMs are also used for local sequence transduction tasks, such as paraphrasing, formality style transfer, Grammatical Error Correction (GEC), and simplification~\cite{kaneko-etal-2022-interpretability,reif-etal-2022-recipe,wu2023chatgpt,wang2022self,kaneko2023controlled}, where only a small portion of the source text is edited.
Most tokens in a source text are kept unchanged in these tasks.
For example, the source text, \textit{``Many years ago, the situation \underline{is} different,''} and the target text, \textit{``Many years ago, the situation \underline{was} different,''} of the GEC task mostly share the common tokens except for the underlined tokens (\textit{is} and \textit{was}).

Existing methods of downstream tasks do not make use of the characteristics of local sequence transduction~\cite{reif-etal-2022-recipe,wu2023chatgpt,wang2022self}, simply generating all target tokens.
In this paper, we hypothesize that this treatment is disadvantageous in achieving high performance in terms of task accuracy and computational time.
More specifically, it is inefficient to generate unchanged tokens (e.g. \textit{Many, years, ago, the, situation, different}) in the previous example because the model must copy many source tokens only to increase the length of the target sequence.

\begin{figure*}[!t]
  \centering
  \includegraphics[width=1\textwidth]{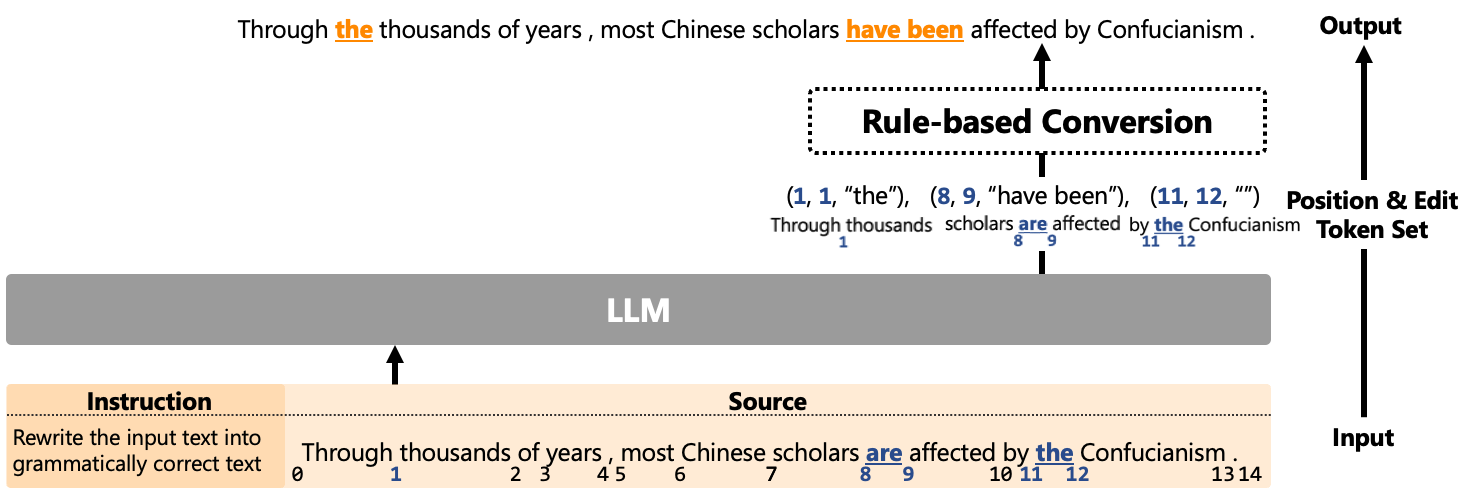}
  \caption{Inference of instruction tuned LLMs using edit spans. LLMs take instruction text and source text as input and output only the positions and tokens for rewriting. Rule-based conversion applies the outputted positions and tokens of the rewriting to the source text and produces the plaintext output.}
  \label{fig:abst}
\end{figure*}

This study proposes to predict a set of edit spans, which represent the changed parts of the target text relative to the source tokens.
Omitting unedited tokens that occupy most of the target text, we can reduce the length of the target text and the inference time for local sequence transduction tasks.
\autoref{fig:abst} shows the process of creating a set of edit spans from source and target texts in GEC. First, we align tokens in the source and target texts to extract the edit locations and tokens and convert them into a set of edit spans. 
In the example shown in \autoref{fig:abst}, the edit spans (1, 1, ``\textit{the}''), (8, 9, ``\textit{have been}''), (12, 13, ``'') are created from the source text \textit{``Through thousands of years, most Chinese scholars \underline{are} greatly affected by \underline{the} Confucianism.''} and the target text \textit{``Through \underline{the} thousands of years, most Chinese scholars \underline{have been} greatly affected by Confucianism.''}.
LLMs are fine-tuned using pairs of source text and edit spans with the instructions. 

We conducted experiments on four local sequence transduction tasks: paraphrasing, formality style transfer, GEC, and simplification.
The proposed method achieved comparable performance to the baseline that directly outputs the target text.
In these tasks, the proposed method could reduce the sequence length on the target side by 32\% on average and by as small as 21\% in GEC.
Furthermore, the proposed method with task-specific fine-tuning achieved state-of-the-art (SoTA) performance in the four tasks.

\section{Edit Spans}

\subsection{Edit Span Extraction}

To extract the editing locations and results of the source and target texts, we calculate the alignment between the tokens in each text.
We use linguistical alignment, which incorporates linguistic information, to perform the alignment~\cite{felice-etal-2016-automatic}.
Linguistical alignment is a method based on the Damerau-Levenshtein algorithm that aligns tokens by considering not only the distance between tokens but also the match of their lemma, part-of-speech, and character features, weighted accordingly.
Taking into account the linguistic information of tokens, linguistical alignment is more accurate compared to alignment methods that only use surface information.
Furthermore, linguistic alignment merges token alignments using recursive rules to create alignments for multiple tokens, such as \textit{``have been''} in \autoref{fig:abst}.

To indicate the edit position identified by the alignment, a 0 is assigned before the first token of the source text, and an index is sequentially assigned to the space after each token.
When the length of the source text is $N$, $N$ is assigned after the last token.
The edit span is represented by the tuple of the start position of the source text, the end position of the source text, and the token result after being edited.

There are three types of edit operations: insert, replace, and delete; we explain them using the example in \autoref{fig:abst}.
The tuple (1, 1, ``\textit{the}'') represents the operation to insert ``\textit{the}''.
In an insertion operation, both the start and end positions are set to the same position where a token is inserted in the source.
The tuple stands for inserting ``\textit{the}'' between the tokens located at the 1st position.
The tuple (8, 9, ``\textit{have been}'') presents the operation to replace ``\textit{are}'' with ``\textit{have been}''.
By specifying the 8th and 9th positions of the source text, this tuple targets the ``\textit{are}'' and rewrites them as ``\textit{have been}''.
The tuple (12, 13, ``'') represents the operation to delete ``\textit{the}''.
It points ``\textit{the}'' by specifying the 12th and 13th positions in the source text.
Because the target token after this edit operation is empty, this tuple corresponds to removing ``\textit{the}''.



\subsection{Instruction Tuning with Edit Spans}

Instruction tuning fine-tunes LLMs by using natural language instructions describing a task~\cite{wei2021finetuned}.
Compared to the conventional fine-tuning that specializes the model for a specific task, instruction tuning aims for generalization to various tasks by training LLMs to respond well to many kinds of instructions.
Therefore, instruction tuning is used for training many LLMs in an open-ended setting~\cite{ouyang2022training,chung2022scaling,wang2022self,wu2023lamini}.
We use the description of local sequence transduction tasks as instructions to perform instruction tuning of LLMs.
We provide the LLMs with instructions and source text, and train the LLMs to generate edit spans.
When there are multiple edits, they are concatenated with commas like \textit{``1 1 the, 8 9 have been, 12 13''}.
When no editing is required in the source text, \textit{``None''} is given as the gold text.

Recent LLMs are expected to have the ability to handle unknown tasks and various tasks, to achieve generality.
It is important that learning through edit spans does not degrade the performance of tasks other than local sequence transduction tasks.
Therefore, we add edit span data to the existing training data for instruction tuning, which includes various tasks, and fine-tune LLMs.

\subsection{Conversion from Edit Spans to Output Text}

To convert the edit spans output by LLMs into plaintext, we use a rule-based approach.
If LLMs generate \textit{``None''}, we use the source text as the final output text.
Otherwise, we split the edit spans by commas and extract the edits.
From each edit, we extract the starting position, ending position, and edited result.
If LLMs generate edits in an incorrect format that do not include start or end positions or edits where the start or end positions exceed the source text range, we ignore them.
To ensure that the token indices do not shift, we apply the edits to the source text in descending order of starting positions.
This conversion is implemented by simple rules with a minimal computational cost.

\section{Experiment Setting}

\subsection{Local Sequence Transduction Taskes}

We conducted experiments on local sequence transduction tasks such as GEC, paraphrasing, formality style transfer, and simplification.

\paragraph{GEC}

We used NUCLE as the training data, CoNLL2013~\cite{ng-etal-2013-conll} as the development data, and CoNLL2014~\cite{ng-etal-2014-conll} as the evaluation data.
The dataset is comprised of essays composed by college students from the National University of Singapore, covering a broad spectrum of subjects, including environmental pollution, healthcare, and more.
We used the M$^{\rm 2}$ score~\cite{dahlmeier-ng-2012-better} as the evaluation metric.
For GEC, we provide the instruction text \textit{``Rewrite the input text into grammatically correct text.''}.

\paragraph{Paraphrasing}

Quora published a dataset that includes more than 400K lines of potential question duplicate pairs\footnote{\url{https://www.kaggle.com/c/quora-question-pairs}}.
Of these pairs, 150K question pairs were labeled as paraphrases.
Only those labeled paraphrase question pairs are used as training, development, and test sets.
We used BLEU-4~\cite{papineni2002bleu}, ROUGE-1, and ROUGE-2~\cite{lin-2004-rouge} to evaluate LLMs, following previous research~\cite{10.1162/tacl_a_00318,meng-etal-2021-conrpg,li-etal-2022-learning-adapt}.
For paraphrasing, we provide the instruction text \textit{``Rewrite the input text into paraphrased text.''}

\paragraph{Style transfer}

We used FST benchmark Grammarly Yahoo Answers Corpus (GYAFC)~\cite{rao-tetreault-2018-dear} for formality style transfer.
GYAFC is a plain corpus that contains pairs of informal and formal sentences conveying the same meaning.
It covers domains such as Entertainment \& Music (E\&M) and Family \& Relationship (F\&R).
We utilized the corpus BLEU in NLTK~\cite{bird-loper-2004-nltk} as described in~\citet{chawla-yang-2020-semi}.
For formality style transfer, we provide the instruction text \textit{``Rewrite the input text into formal text.''}

\paragraph{Simplification}

We used WikiSmall\footnote{\url{https://github.com/XingxingZhang/dress}}~\cite{zhu-etal-2010-monolingual,zhang-lapata-2017-sentence} as the training data and ASSET~\cite{alva-manchego-etal-2020-asset} and TurkCorpus~\cite{xu-etal-2016-optimizing} as the evaluation data.
We used SARI~\cite{xu-etal-2016-optimizing} to evaluate LLMs, which compares the generated text with the target text and calculates the average F1 score for addition, keep, and deletion operations.
For text simplification, we provide the instruction text \textit{``Rewrite the input text into simpler text.''}

\subsection{Open-ended Tasks}

The rules of edit spans differ from the rules in the raw text, which could potentially have a negative impact on the performance of tasks other than local sequence transduction.
By combining open-ended instruction tuning data and edit spans instruction tuning data, we can train LLMs and investigate their impact on other tasks as well.

We utilize the databricks-dolly-15k dataset\footnote{\url{https://huggingface.co/datasets/databricks/databricks-dolly-15k/viewer/databricks--databricks-dolly-15k}} by randomly dividing it into 13K for training, 1K for development, and 1K for evaluation.
databricks-dolly-15k is a publicly available dataset consisting of instructional records created by numerous Databricks employees.
It covers various behavioral categories described in InstructGPT~\cite{ouyang2022training}, such as brainstorming, classification, closed QA, generation, information extraction, open QA, and summarization.
We sampled 3K instances for each of the tasks: GEC, paraphrasing, style transfer, and simplification, resulting in a total of 12K instruction instances.
We fine-tuned LLMs using a combined dataset of all these instructions, totaling 25K instances.

We used BERTScore\footnote{\url{https://github.com/Tiiiger/bert_score}}~\cite{zhang2019bertscore} as our evaluation metric.
BERTScore is an evaluation method that measures the similarity between generated text and target text using contextual embeddings from pre-trained models.
We utilized RoBERTa~\cite{liu2019roberta} (roberta-large\footnote{\url{https://huggingface.co/roberta-large}})  as the BERTScore models.

\subsection{Instruction Tuning Settings}
\label{sec:instruction}

We used the following four LLMs for our experiments: MPT (mpt-7b)\footnote{\url{https://huggingface.co/mosaicml/mpt-7b}}~\cite{MosaicML2023Introducing}, OPT (opt-6.7b)\footnote{\url{https://huggingface.co/facebook/opt-6.7b}}~\cite{zhang2022opt}, LLaMA (llama-7b)\footnote{\url{https://github.com/facebookresearch/llama}}~\cite{touvron2023llama}, and BLOOM (bloom-7b1)\footnote{\url{https://huggingface.co/bigscience/bloom-7b1}}~\cite{scao2022bloom}.

We used the code for instruction tuning from Stanford Alpaca~\cite{alpaca} code\footnote{\url{https://github.com/tatsu-lab/stanford_alpaca}} for instruction tuning.
We set the number of epochs to 3 and used a batch size of 32.
The learning rate was set to 2e-5, with a warmup rate of 0.03, and we employed a cosine learning rate schedule.
These hyperparameters were determined following Stanford Alpaca.
We report the average results of three models trained with different seeds for instruction tuning.
We used four nodes, each containing eight NVIDIA A100 GPUs.
We used the code\footnote{\url{https://github.com/chrisjbryant/errant}} for linguistical alignment provided by \citet{felice-etal-2016-automatic}.

\paragraph{Baselines}

We compare the results of the proposed method with the results of LLMs fine-tuned for instruction tuning using the target text as the ground truth instead of using edit spans.
This comparison examines whether edit spans can reduce computational costs during inference without compromising performance.

\section{Experiment}

\subsection{Performance on Local Sequence Transduction Tasks}

\begin{table*}[t]
\small
\centering
\begin{tabular}{llcccc}
\toprule
& & GEC & Paraphrasing & Style transfer & Simplification  \\
\midrule
\multirow{4}{*}{Plain}& MPT & 68.0 & 37.9/66.5/47.1 & 78.9/81.2 & 46.3/41.1 \\
& OPT & 65.7 & 35.2/63.2/45.4 & 75.0/77.2 & 43.7/40.5 \\
& LLaMA & 68.2 & \textbf{39.3}/69.0/47.2 & \textbf{79.5}/81.0 & 48.0/41.9 \\
& BLOOM & 66.4 & 37.0/66.4/46.1 & 78.2/79.9 & 45.0/41.0 \\
\midrule
\multirow{4}{*}{Edit spans}& MPT & \underline{68.5} & \underline{38.2}/\underline{66.7}/47.1 & 78.2/\underline{81.3} & \underline{46.6}/\underline{41.3}  \\
& OPT & \underline{66.2} & 34.1/61.2/43.9 & \underline{75.6}/\underline{77.9} & \underline{43.9}/40.3 \\
& LLaMA & \underline{\textbf{69.1}} & 39.0/\underline{\textbf{69.2}}/\underline{\textbf{47.6}} & 79.3/\underline{\textbf{81.2}} & \underline{\textbf{48.3}}/\underline{\textbf{42.0}} \\
& BLOOM & 65.8 & \underline{37.2}/66.1/\underline{46.3} & 78.0/\underline{80.3} & 44.8/40.7 \\
\bottomrule
\end{tabular}
\caption{The performance of four LLMs fine-tuned with edit spans and plain data instructions on four local sequence transduction tasks. The \textbf{bold} values indicate the highest performance for each task. The \underline{underlined} values indicate when edit spans exceed the baseline.}
\label{tbl:performance}
\end{table*}

To demonstrate the contribution of edit spans to performance improvement, we first compare the baseline performance with fine-tuned data using plain text.
\autoref{tbl:performance} shows the results of performance comparison between the baseline and the proposed method in the GEC, paraphrasing, style transfer, and simplification tasks.
Out of 32 cases, performance improvement was observed in 19 cases, and edit spans contributed to the performance enhancement.
Furthermore, it can be observed that the LLaMA trained with edit spans achieves the highest performance in most cases.

\subsection{Reducing Text Length}

\begin{figure*}[t]
\begin{tabular}{cc}
\begin{minipage}[t]{0.45\hsize}
\centering
\includegraphics[keepaspectratio, scale=0.5]{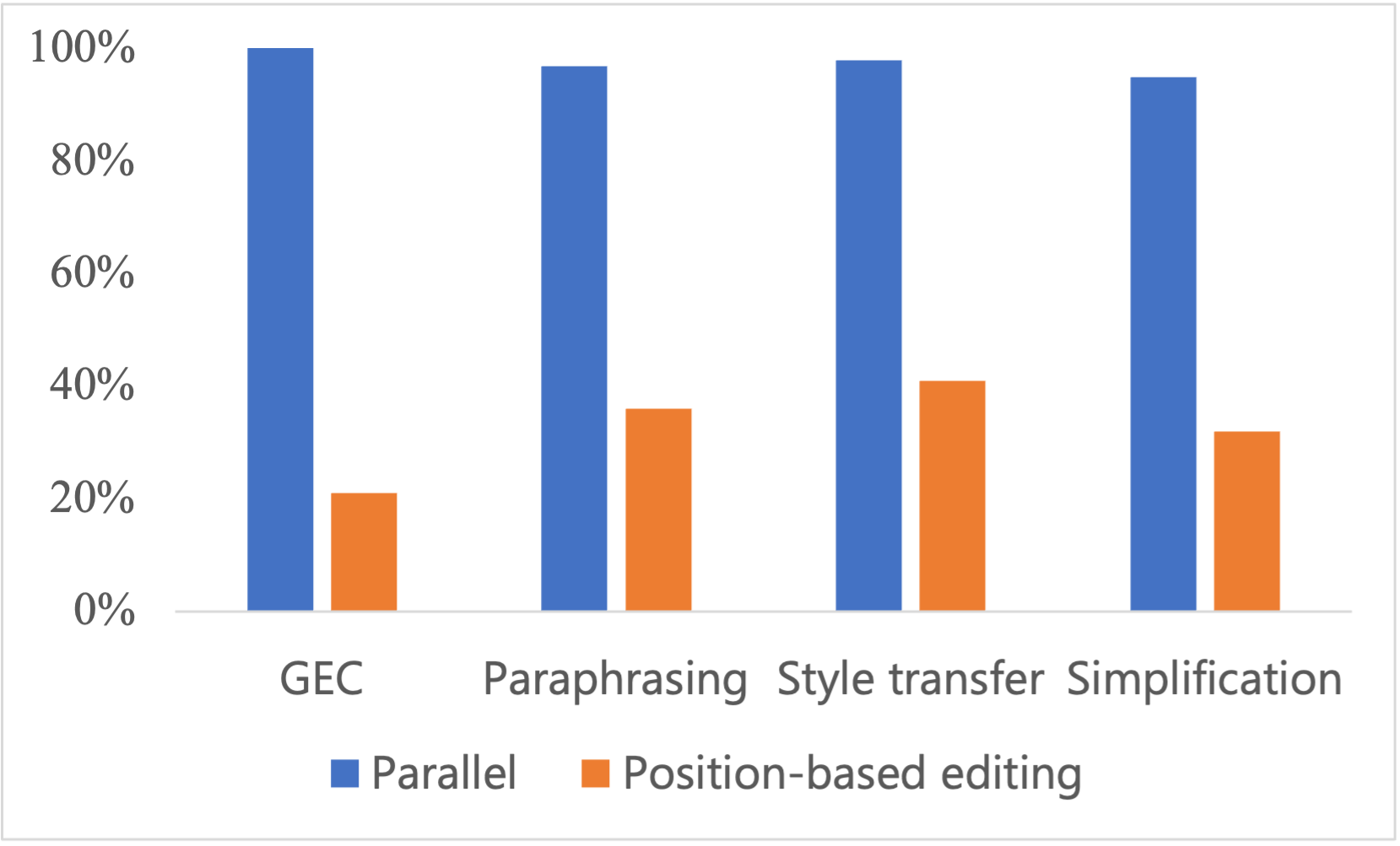}
\subcaption{MPT}
\label{fig:mpt}
\end{minipage} &
\begin{minipage}[t]{0.45\hsize}
\centering
\includegraphics[keepaspectratio, scale=0.5]{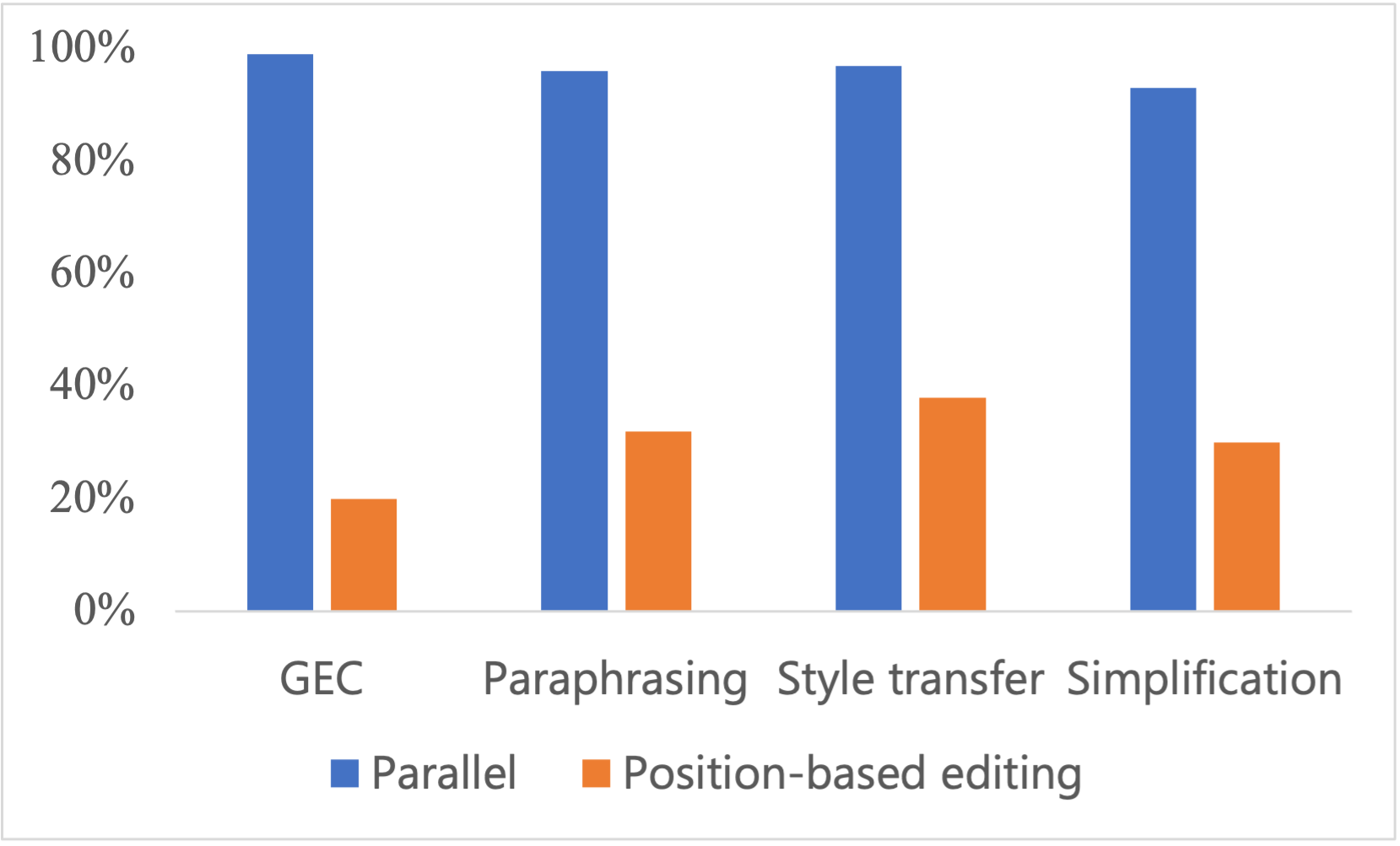}
\subcaption{OPT}
\label{fig:opt}
\end{minipage} \\[2.5em]

\begin{minipage}[t]{0.45\hsize}
\centering
\includegraphics[keepaspectratio, scale=0.5]{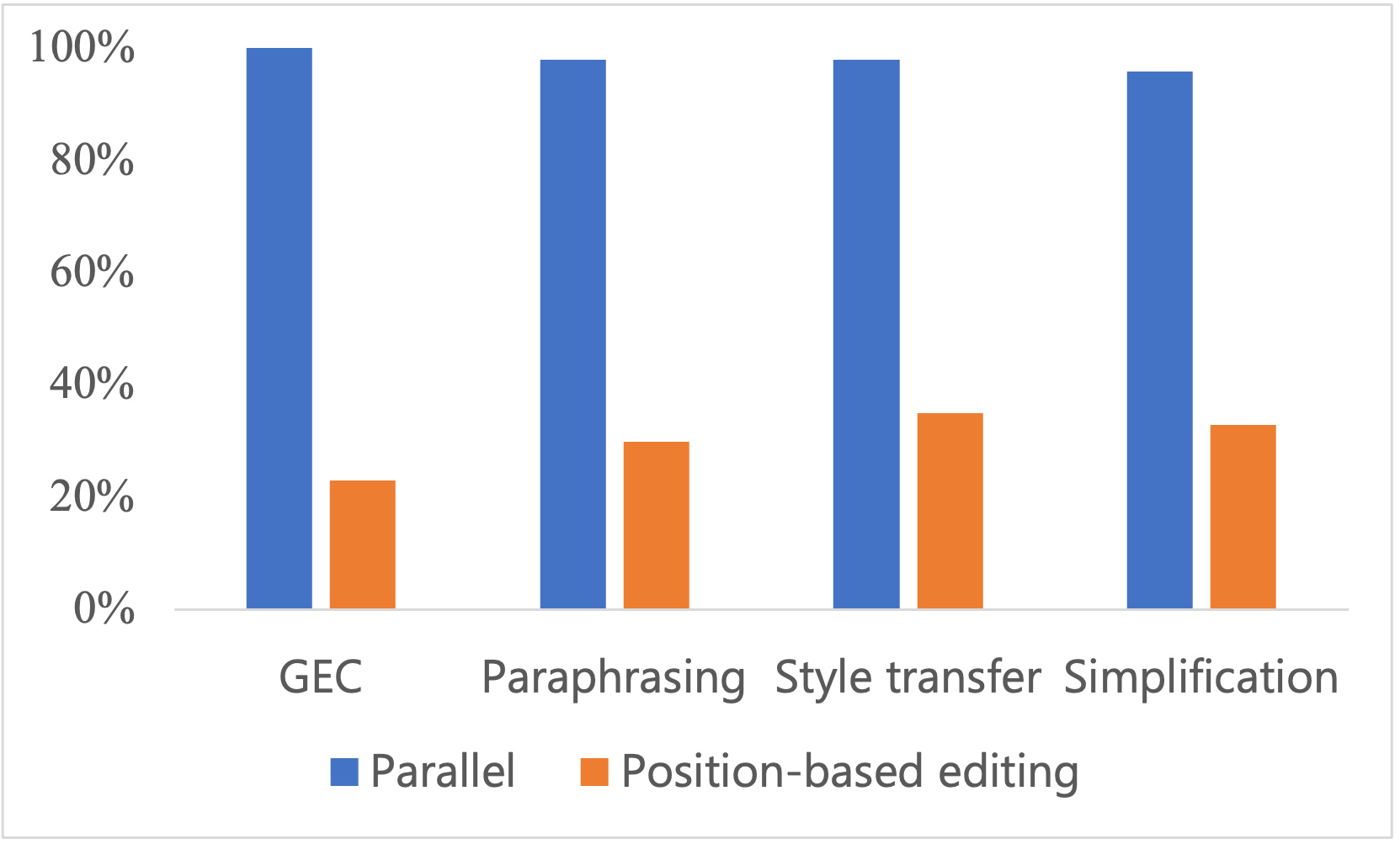}
\subcaption{LLaMA}
\label{fig:llama}
\end{minipage} &
\begin{minipage}[t]{0.45\hsize}
\centering
\includegraphics[keepaspectratio, scale=0.5]{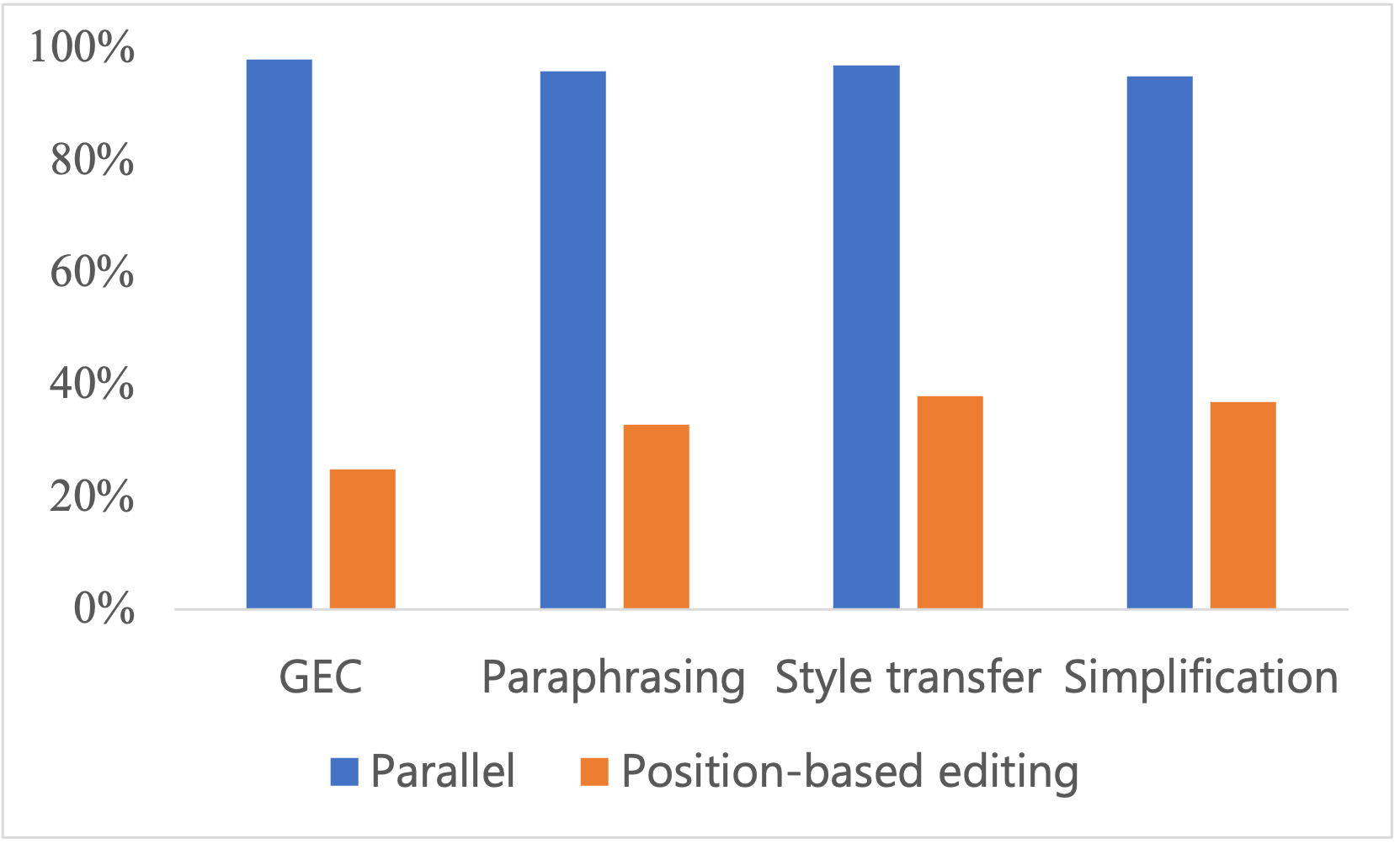}
\subcaption{BLOOM}
\label{fig:bloom}
\end{minipage} 
\end{tabular}
\caption{The ratio of output text length to target text length when MPT, OPT, LLaMA, and BLOOM are fine-tuned with plain data and edit span data, respectively.}
\label{fig:compression}
\end{figure*}

We examine how much the fine-tuning of LLMs with edit span data reduced the length of the output text.
\autoref{fig:compression} shows the ratio of output text length to target text length when fine-tuned with plain data and edit span data, respectively, on the development data for each task.
The proposed method successfully compresses the output text across all tasks, independent of the model used;
it achieves text compression in the range of 21\% in the most compressed cases and 41\% even in the least compressed cases.
In GEC, there are cases where grammatically correct text is provided as source text.
In such cases, the model does not need to make any revisions and can simply output \textit{``None''}, resulting in significant compression in GEC.

\subsection{Performance on Open-ended Task}

In open-ended tasks, the target texts are written in plain text, while edit spans introduce significant differences in text formatting. This misalignment in text representation may potentially impact the performance of open-ended tasks.
Therefore, we aim to demonstrate that edit spans do not significantly degrade the performance of open-ended tasks.

\autoref{tbl:open} shows the scores for each LLM when using RoBERTa as BERTScore models on the 1K subset of the databricks-dolly-15k dataset, which was divided for evaluation.
This indicates that the proposed method achieves efficient computational cost during inference without significantly sacrificing open-ended task performance.

To maintain performance in open-ended tasks, the proposed method combines data from both local sequence transduction tasks and open-ended tasks.
To demonstrate the effectiveness of combining open-ended task data, we also investigate the open-ended task performance of instruction-tuned LLMs when solely trained on local sequence transduction task data.

\autoref{tbl:only} demonstrates the performance difference on the 1K split of the databricks-dolly-15k dataset, evaluating LLMs trained on both open-ended task and local sequence transduction task data versus LLMs trained solely on local sequence transduction task data.
The performance decreases when not using open-ended task data for training, both in terms of plain text and edit spans.
This is likely because open-ended task data consists of plain text, while edit spans include totally different text formats, leading to a larger disparity in task requirements.

\begin{table}[t]
\small
\centering
\begin{tabular}{lcccc}
\toprule
& & BERTScore   \\
\midrule
\multirow{4}{*}{Plain}& MPT & 81.5  \\
& OPT & 79.3  \\
& LLaMA & 81.8 \\
& BLOOM & 79.9 \\
\midrule
\multirow{4}{*}{Edit span}& MPT & 81.0 \\
& OPT & 78.6 \\
& LLaMA & 81.3 \\
& BLOOM & 79.5 \\
\bottomrule
\end{tabular}
\caption{Scores using BERTScore on the databricks-dolly-15k dataset, which was divided for evaluation.}
\label{tbl:open}
\end{table}

\begin{table}[t]
\small
\centering
\begin{tabular}{lcccc}
\toprule
& & BERTScore diff.  \\
\midrule
\multirow{4}{*}{Plain}& MPT & -5.2 \\
& OPT & -5.7 \\
& LLaMA & -4.4 \\
& BLOOM & -6.2 \\
\midrule
\multirow{4}{*}{Edit span}& MPT & -8.1 \\
& OPT & -8.6 \\
& LLaMA & -6.9 \\
& BLOOM & -7.6 \\
\bottomrule
\end{tabular}
\caption{The performance difference between instruction tuned LLMs using local sequence transduction task and open-ended task datasets, and instruction tuned LLMs using only local sequence transduction task datasets.}
\label{tbl:only}
\end{table}

\subsection{The Accuracy of Edits Generated by the LLMs}

Even if the edit span text is different, there are cases where the text is transformed by the rule, and the text matches.
For example, in GEC, the model is given the input \textit{``This technology could also be seen as invasion of human privacy.''} and the model outputs \textit{``7 9 invading''}.
In this case, even with the alternate  edit span text \textit{``7 8 invading, 8 9''}, the conversion based on the rules would result in the same output text.
However, this would increase the sentence length by the index, creating room for improvement in terms of computational cost.
Therefore, we investigate how well edit span of the model matches the results using linguistic alignment.

First, we convert the edit spans generated by the model to plain text using rules.
From the converted plain text and the source text, we create edit spans using linguistic alignment and calculate the percentage of agreement with the edit spans output by the model.
Only when the start position $s$, end position $e$, and the edit token $r$ all match exactly is it considered a correct answer.

\autoref{tbl:acc} shows the percentage of agreement between the edit spans output by the LLMs and the edit spans extracted by the linguistic alignment in the development data for each task.
The proposed method achieves more than 90\% agreement in 13 out of 16 settings.
This indicates that LLMs are able to learn the extraction rules for linguistical alignment through instruction tuning.

\begin{table*}[t]
\small
\centering
\begin{tabular}{llcccc}
\toprule
& GEC & Paraphrasing & Style transfer & Simplification  \\
\midrule
MPT & 96.6 & 95.0 & 89.2 & 94.7 \\
OPT & 93.3 & 91.9 & 88.8 & 92.7 \\
LLaMA & 99.0 & 96.2 & 92.6 & 95.4 \\
BLOOM & 94.2 & 92.5 & 89.4 & 93.5 \\
\bottomrule
\end{tabular}
\caption{Agreement between edit spans generated by LLMs and edit spans extracted by linguistic alignment.}
\label{tbl:acc}
\end{table*}

\subsection{Task-specific Fine-tuning}

In the previous experiments, LLMs were trained by combining data from the four local sequence transduction tasks and the open-ended task.
To explore the maximum potential performance of the proposed method, we fine-tune LLMs with task-specific focus using edit span data.
We fine-tune LLMs for each task using all available training data.
In this case, we specialize LLMs for specific tasks without the need for instruction texts.
Therefore, we trained the LLMs by providing only the source texts as input.

We trained LLaMA, which showed the highest performance in the local sequence transduction tasks.
We set the number of epochs to 2 and used a batch size of 32.
The learning rate was set to 1e-5, with a warmup rate of 0.03, and we employed a cosine learning rate schedule.
Following the exploration method described in Section \ref{sec:instruction}, we determined the hyperparameters for our experiments.

\autoref{tbl:task} shows the results of performance comparison with existing studies on GEC, paraphrasing, style transfer, and simplification tasks.
The proposed method outperforms existing studies by 1.8 points in GEC, 0.9, 1.2, and 2.3 points in paraphrasing, 1.9 and 1.3 points in style transfer, and 1.2 and 0.7 points in simplification tasks, respectively.
Thus, the proposed method achieves the SoTA performance in all tasks.
From these results, it can be concluded that edit spans are an effective method, even in task-specific fine-tuning scenarios.

\begin{table*}[t]
\centering
\small
\begin{subtable}{0.4\linewidth}
\centering
\begin{tabular}{lc}
\toprule
& GEC  \\
\midrule
\cite{kaneko-etal-2020-encoder} & 65.2 \\
\cite{omelianchuk-etal-2020-gector} & 66.5 \\
\cite{qorib-etal-2022-frustratingly} & 69.5 \\
Edit span & \textbf{71.3} \\
\bottomrule
\end{tabular}
\caption{M$^2$ scores on the CoNLL2014 dataset.}
\label{tab:gec}
\end{subtable}%
\hfill
\begin{subtable}{0.4\linewidth}
\centering
\begin{tabular}{lc}
\toprule
& Paraphrasing  \\
\midrule
\cite{10.1162/tacl_a_00318} & 38.0/68.1/45.7 \\
\cite{meng-etal-2021-conrpg} & 26.8/65.0/38.5 \\
\cite{li-etal-2022-learning-adapt} & 39.3/70.8/48.3 \\
Edit span & \textbf{41.2/72.0/50.6} \\
\bottomrule
\end{tabular}
\caption{BLEU-4, ROUGE-1, and ROUGE-2 scores on the Quora dataset.}
\label{tab:paraphrasing}
\end{subtable}\\[1em]

\begin{subtable}{0.4\linewidth}
\centering
\begin{tabular}{lc}
\toprule
& Style transfer  \\
\midrule
\cite{chawla-yang-2020-semi} & 76.2/79.9 \\
\cite{lai-etal-2021-thank} & 76.5/79.3 \\
\cite{liu-etal-2022-semi} & 78.8/81.4 \\
Edit span & \textbf{80.7/82.7} \\
\bottomrule
\end{tabular}
\caption{NLTK BLEU scores on the E\&M and F\&R datasets.}
\label{tab:style}
\end{subtable}%
\hfill
\begin{subtable}{0.4\linewidth}
\centering
\begin{tabular}{lc}
\toprule
& Simplification  \\
\midrule
\cite{martin-etal-2020-controllable} & 40.1/41.4 \\
\cite{martin-etal-2022-muss} & 44.2/42.6 \\
\cite{feng2023sentence} & 47.9/41.8 \\
Edit span & \textbf{49.1/43.5} \\
\bottomrule
\end{tabular}
\caption{SARI scores on ASSET and TurkCorpus datasets.}
\label{tab:simplification}
\end{subtable}
\caption{Performance comparison with previous studies on GEC, paraphrasing, style transfer, and simplification tasks.}
\label{tbl:task}
\end{table*}

\subsection{Example of LLMs Output Using Edit Spans}

\begin{table*}[t]
\small
\centering
\begin{tabular}{llcccc}
\toprule
Source text & Since we do not to bring cash to pay for the transportation fee , enormous time has been saved \\
& for everybody . \\
Target text & Since we do not \underline{need} to bring cash to pay for the transportation fee , enormous time has been saved \\
& for everybody . \\
Target edit span & 4 4 need \\
\midrule
Plain & Since we do not to bring cash to pay for the transportation fee , enormous time has been saved \\
& for everybody . \\
System edit span & 4 4 need \\
\bottomrule
\end{tabular}
\caption{Outputs in plain text and edit span formats respectively by LLaMA in the CoNLL2013.}
\label{tbl:example}
\end{table*}

\autoref{tbl:example} shows the output in CoNLL2013 for LLaMA using edit span and LLaMA outputting plain text.
The normal model outputting plain text outputs 23 tokens, while the model using edit span outputs only 3 tokens.
The output of the model using the edit span is a much shorter sequence than the original model that outputs plain text.
Furthermore, LLaMA, which outputs in plain text, is unable to correct a grammatical error.
In a local sequence transduction task, most tokens in the source text and target text are common, and the model tends to learn just to copy the input tokens~\cite{rastogi-etal-2016-weighting}.
Contrarily, our model that uses edit spans outputs only the edited parts. 
Thus simply copying the input is not an issue for our model.

\section{Related Work}

\subsection{Efficient LLMs}

Most of the methods for achieving efficient LLMs involve improving the memory complexity of self-attention mechanisms or enhancing the overall efficiency of the Transformer architecture~\cite{tay2022efficient,loem2022neighbors}.
In the initial stages, the modifications made to self-attention focused on reducing the computational complexity by introducing sparsity in the attention matrix.
This was accomplished by restricting the attention's scope to predetermined patterns, such as local windows and fixed stride block patterns~\cite{liu2018generating,qiu-etal-2020-blockwise,beltagy2020longformer}.
A natural extension to the blockwise method is to connect these blocks via recurrence.
\citet{dai-etal-2019-transformer} introduced a mechanism of segment-level recurrence that establishes connections among multiple segments and blocks.

An expansion upon fixed, predetermined patterns is the utilization of learnable patterns.
Models that incorporate learnable patterns aim to acquire the access pattern through data-driven methods.
One crucial aspect of learning patterns is to establish a concept of token relevance and subsequently assign tokens to buckets or clusters~\cite {vyas2020fast,wang-etal-2021-cluster,kitaev2020reformer,tay2020sparse,roy-etal-2021-efficient}.

Another approach is to utilize a trainable side memory module capable of accessing multiple tokens simultaneously~\cite{sukhbaatar2019augmenting,ainslie-etal-2020-etc,beltagy2020longformer}.
A prevalent example is the global neural memory, which can access the entire sequence.
The global tokens function as a type of model memory, learning to gather information from the input sequence tokens.

Another method to enhance efficiency is by utilizing low-rank approximations of the self-attention matrix to improve computational performance~\cite{wang2020linformer}, and to view the attention mechanism through kernelization~\cite{choromanski2020masked,peng2021random}.
Sparse models selectively activate a fraction of the parameters, resulting in an improved parameter to FLOPs ratio in general~\cite{fedus2022switch}.

As a way to reduce the length of the text, \citet{cheng2023batch} proposed including multiple examples in one prompt and inferring in parallel.

These techniques, unlike our research, do not alter the writing style of the target text, and edit spans can be used in conjunction with these methods.

\subsection{Edit-based Model}

Since the question of necessarily using the seq2seq model for local sequence transduction tasks was raised~\cite{rastogi-etal-2016-weighting,schnober-etal-2016-still}, various edit-based models have been proposed.
\citet{guu-etal-2018-generating} proposed a language model that initially selects a prototype sentence from the training dataset and subsequently modifies it to create a new sentence.
\citet{ribeiro-etal-2018-local} introduced a method for representing general string transduction problems as sequence labeling.
\citet{koide2018neural} proposed the model implemented to analyze the evolution of biological sequences driven by substitution, insertion, and deletion edit operations, achieving improved accuracy on protein secondary structure prediction.
\citet{awasthi-etal-2019-parallel} presented a parallel iterative edit model reducing decoding time for local sequence transduction tasks.
\citet{gu2019levenshtein} developed the Levenshtein Transformer, a non-autoregressive model using edit operations.
\cite{mallinson-etal-2020-felix} introduced FELIX, an adaptable text-editing approach for the generation that aims to leverage the advantages of decoding with bi-directional contexts and self-supervised pretraining to the fullest extent.
\cite{xu2021editor} presented an Edit-Based Transformer with Repositioning, which enhances sequence generation flexibility by seamlessly incorporating user-specified preferences in output lexical choice.
\citet{reid-neubig-2022-learning} proposed the modeling of editing processes, encompassing the iterative generation of sequences as a whole.
They establish a conceptual framework to explain the probability of multi-step edits and outline neural models capable of learning a generative model of sequences by leveraging these multi-step edits.

However, these methods have different architectures from LLMs.
Therefore, it is not easy to apply them to LLMs, unlike our method, which can train models by simply changing the output text.

\subsection{LLMs for Local Sequence Transduction Tasks}

In GEC, the model based on GPT-3 achieves state-of-the-art in unsupervised settings~\cite{loem2023exploring}.
\citet{fang2023chatgpt} showed that ChatGPT corrects input text very fluently.
\citet{yamashita-etal-2020-cross,rothe-etal-2021-simple,sun2022unified} proposed a method for multilingual GEC using multilingual LLMs.
\citet{Feng2023SentenceSV} investigated the performance of few-shot and zero-shot of GPT3 and ChatGPT in the simplification.
\citet{anschutz2023language} used LLMs for German simplification and found them to be effective in languages with little parallel data.
\cite{witteveen-andrews-2019-paraphrasing} verified the performance of GPT-2~\cite{radford2019language} in paraphrasing.
\citet{wahle-etal-2022-large} investigated the utilization of T5 and GPT3 in generating machine-generated paraphrases for scientific articles sourced from arXiv, student theses, and Wikipedia.
\citet{reif-etal-2022-recipe} introduced a method based on GPT-3 that solely relies on natural language instruction and does not necessitate model fine-tuning or exemplars in the desired style.
\citet{malmi-etal-2020-unsupervised} proposed a method of using LLMs for style transfer where no parallel data is available.
On the other hand, these studies did not target the efficiency of LLMs based on the edit.

\section{Conclusion}

In this study, we proposed to predict a set of edit spans, which represent the changed parts of the target text relative to the source tokens.
We showed our method omits unedited tokens that occupy most of the target text and reduces the length of the target text and the inference time for local sequence transduction tasks.
Moreover, we reported that instruction tuning with the proposed method achieves state-of-the-art performance in the four tasks.


\section*{Limitations}

In our preliminary experiments, even high-performance LLMs such as GPT-3~\cite{NEURIPS2020_1457c0d6} and ChatGPT~\cite{chatgpt} could not generate edit spans with zero-shot and few-shot.
In particular, indexes could not be generated correctly.
Therefore, it is a future work to apply the proposed method to zero-shot and few-shot.
Moreover, the use of edit span is not necessarily effective for tasks, such as machine translation and dialogue, other than the local sequence transduction task, where many tokens in the source and target texts are not common.

\section*{Acknowledgements}
These research results were obtained from the commissioned research (No.225) by National Institute of Information and Communications Technology (NICT), Japan.

\bibliography{emnlp2023}

\begin{thebibliography}{87}
\expandafter\ifx\csname natexlab\endcsname\relax\def\natexlab#1{#1}\fi

\bibitem[{Ainslie et~al.(2020)Ainslie, Ontanon, Alberti, Cvicek, Fisher, Pham,
  Ravula, Sanghai, Wang, and Yang}]{ainslie-etal-2020-etc}
Joshua Ainslie, Santiago Ontanon, Chris Alberti, Vaclav Cvicek, Zachary Fisher,
  Philip Pham, Anirudh Ravula, Sumit Sanghai, Qifan Wang, and Li~Yang. 2020.
\newblock \href {https://doi.org/10.18653/v1/2020.emnlp-main.19} {{ETC}:
  Encoding long and structured inputs in transformers}.
\newblock In \emph{Proceedings of the 2020 Conference on Empirical Methods in
  Natural Language Processing (EMNLP)}, pages 268--284, Online. Association for
  Computational Linguistics.

\bibitem[{Alva-Manchego et~al.(2020)Alva-Manchego, Martin, Bordes, Scarton,
  Sagot, and Specia}]{alva-manchego-etal-2020-asset}
Fernando Alva-Manchego, Louis Martin, Antoine Bordes, Carolina Scarton,
  Beno{\^\i}t Sagot, and Lucia Specia. 2020.
\newblock \href {https://doi.org/10.18653/v1/2020.acl-main.424} {{ASSET}: {A}
  dataset for tuning and evaluation of sentence simplification models with
  multiple rewriting transformations}.
\newblock In \emph{Proceedings of the 58th Annual Meeting of the Association
  for Computational Linguistics}, pages 4668--4679, Online. Association for
  Computational Linguistics.

\bibitem[{Ansch{\"u}tz et~al.(2023)Ansch{\"u}tz, Oehms, Wimmer, Jezierski, and
  Groh}]{anschutz2023language}
Miriam Ansch{\"u}tz, Joshua Oehms, Thomas Wimmer, Bart{\l}omiej Jezierski, and
  Georg Groh. 2023.
\newblock Language models for german text simplification: Overcoming parallel
  data scarcity through style-specific pre-training.
\newblock \emph{arXiv preprint arXiv:2305.12908}.

\bibitem[{Awasthi et~al.(2019)Awasthi, Sarawagi, Goyal, Ghosh, and
  Piratla}]{awasthi-etal-2019-parallel}
Abhijeet Awasthi, Sunita Sarawagi, Rasna Goyal, Sabyasachi Ghosh, and Vihari
  Piratla. 2019.
\newblock \href {https://doi.org/10.18653/v1/D19-1435} {Parallel iterative edit
  models for local sequence transduction}.
\newblock In \emph{Proceedings of the 2019 Conference on Empirical Methods in
  Natural Language Processing and the 9th International Joint Conference on
  Natural Language Processing (EMNLP-IJCNLP)}, pages 4260--4270, Hong Kong,
  China. Association for Computational Linguistics.

\bibitem[{Beltagy et~al.(2020)Beltagy, Peters, and
  Cohan}]{beltagy2020longformer}
Iz~Beltagy, Matthew~E Peters, and Arman Cohan. 2020.
\newblock Longformer: The long-document transformer.
\newblock \emph{arXiv preprint arXiv:2004.05150}.

\bibitem[{Bender et~al.(2021)Bender, Gebru, McMillan-Major, and
  Shmitchell}]{bender2021dangers}
Emily~M Bender, Timnit Gebru, Angelina McMillan-Major, and Shmargaret
  Shmitchell. 2021.
\newblock On the dangers of stochastic parrots: Can language models be too big?
\newblock In \emph{Proceedings of the 2021 ACM conference on fairness,
  accountability, and transparency}, pages 610--623.

\bibitem[{Bird and Loper(2004)}]{bird-loper-2004-nltk}
Steven Bird and Edward Loper. 2004.
\newblock \href {https://aclanthology.org/P04-3031} {{NLTK}: The natural
  language toolkit}.
\newblock In \emph{Proceedings of the {ACL} Interactive Poster and
  Demonstration Sessions}, pages 214--217, Barcelona, Spain. Association for
  Computational Linguistics.

\bibitem[{Brown et~al.(2020)Brown, Mann, Ryder, Subbiah, Kaplan, Dhariwal,
  Neelakantan, Shyam, Sastry, Askell, Agarwal, Herbert-Voss, Krueger, Henighan,
  Child, Ramesh, Ziegler, Wu, Winter, Hesse, Chen, Sigler, Litwin, Gray, Chess,
  Clark, Berner, McCandlish, Radford, Sutskever, and
  Amodei}]{NEURIPS2020_1457c0d6}
Tom Brown, Benjamin Mann, Nick Ryder, Melanie Subbiah, Jared~D Kaplan, Prafulla
  Dhariwal, Arvind Neelakantan, Pranav Shyam, Girish Sastry, Amanda Askell,
  Sandhini Agarwal, Ariel Herbert-Voss, Gretchen Krueger, Tom Henighan, Rewon
  Child, Aditya Ramesh, Daniel Ziegler, Jeffrey Wu, Clemens Winter, Chris
  Hesse, Mark Chen, Eric Sigler, Mateusz Litwin, Scott Gray, Benjamin Chess,
  Jack Clark, Christopher Berner, Sam McCandlish, Alec Radford, Ilya Sutskever,
  and Dario Amodei. 2020.
\newblock Language models are few-shot learners.
\newblock In \emph{Advances in Neural Information Processing Systems},
  volume~33, pages 1877--1901. Curran Associates, Inc.

\bibitem[{Chawla and Yang(2020)}]{chawla-yang-2020-semi}
Kunal Chawla and Diyi Yang. 2020.
\newblock \href {https://doi.org/10.18653/v1/2020.findings-emnlp.212}
  {Semi-supervised formality style transfer using language model discriminator
  and mutual information maximization}.
\newblock In \emph{Findings of the Association for Computational Linguistics:
  EMNLP 2020}, pages 2340--2354, Online. Association for Computational
  Linguistics.

\bibitem[{Cheng et~al.(2023)Cheng, Kasai, and Yu}]{cheng2023batch}
Zhoujun Cheng, Jungo Kasai, and Tao Yu. 2023.
\newblock Batch prompting: Efficient inference with large language model apis.
\newblock \emph{arXiv preprint arXiv:2301.08721}.

\bibitem[{Choromanski et~al.(2020)Choromanski, Likhosherstov, Dohan, Song,
  Gane, Sarlos, Hawkins, Davis, Belanger, Colwell
  et~al.}]{choromanski2020masked}
Krzysztof Choromanski, Valerii Likhosherstov, David Dohan, Xingyou Song,
  Andreea Gane, Tamas Sarlos, Peter Hawkins, Jared Davis, David Belanger, Lucy
  Colwell, et~al. 2020.
\newblock Masked language modeling for proteins via linearly scalable
  long-context transformers.
\newblock \emph{arXiv preprint arXiv:2006.03555}.

\bibitem[{Chowdhery et~al.(2022)Chowdhery, Narang, Devlin, Bosma, Mishra,
  Roberts, Barham, Chung, Sutton, Gehrmann et~al.}]{chowdhery2022palm}
Aakanksha Chowdhery, Sharan Narang, Jacob Devlin, Maarten Bosma, Gaurav Mishra,
  Adam Roberts, Paul Barham, Hyung~Won Chung, Charles Sutton, Sebastian
  Gehrmann, et~al. 2022.
\newblock Palm: Scaling language modeling with pathways.
\newblock \emph{arXiv preprint arXiv:2204.02311}.

\bibitem[{Chung et~al.(2022)Chung, Hou, Longpre, Zoph, Tay, Fedus, Li, Wang,
  Dehghani, Brahma et~al.}]{chung2022scaling}
Hyung~Won Chung, Le~Hou, Shayne Longpre, Barret Zoph, Yi~Tay, William Fedus,
  Eric Li, Xuezhi Wang, Mostafa Dehghani, Siddhartha Brahma, et~al. 2022.
\newblock Scaling instruction-finetuned language models.
\newblock \emph{arXiv preprint arXiv:2210.11416}.

\bibitem[{Dahlmeier and Ng(2012)}]{dahlmeier-ng-2012-better}
Daniel Dahlmeier and Hwee~Tou Ng. 2012.
\newblock \href {https://aclanthology.org/N12-1067} {Better evaluation for
  grammatical error correction}.
\newblock In \emph{Proceedings of the 2012 Conference of the North {A}merican
  Chapter of the Association for Computational Linguistics: Human Language
  Technologies}, pages 568--572, Montr{\'e}al, Canada. Association for
  Computational Linguistics.

\bibitem[{Dai et~al.(2019)Dai, Yang, Yang, Carbonell, Le, and
  Salakhutdinov}]{dai-etal-2019-transformer}
Zihang Dai, Zhilin Yang, Yiming Yang, Jaime Carbonell, Quoc Le, and Ruslan
  Salakhutdinov. 2019.
\newblock \href {https://doi.org/10.18653/v1/P19-1285} {Transformer-{XL}:
  Attentive language models beyond a fixed-length context}.
\newblock In \emph{Proceedings of the 57th Annual Meeting of the Association
  for Computational Linguistics}, pages 2978--2988, Florence, Italy.
  Association for Computational Linguistics.

\bibitem[{Fang et~al.(2023)Fang, Yang, Lan, Wong, Hu, Chao, and
  Zhang}]{fang2023chatgpt}
Tao Fang, Shu Yang, Kaixin Lan, Derek~F Wong, Jinpeng Hu, Lidia~S Chao, and Yue
  Zhang. 2023.
\newblock Is chatgpt a highly fluent grammatical error correction system? a
  comprehensive evaluation.
\newblock \emph{arXiv preprint arXiv:2304.01746}.

\bibitem[{Fedus et~al.(2022)Fedus, Zoph, and Shazeer}]{fedus2022switch}
William Fedus, Barret Zoph, and Noam Shazeer. 2022.
\newblock Switch transformers: Scaling to trillion parameter models with simple
  and efficient sparsity.
\newblock \emph{The Journal of Machine Learning Research}, 23(1):5232--5270.

\bibitem[{Felice et~al.(2016)Felice, Bryant, and
  Briscoe}]{felice-etal-2016-automatic}
Mariano Felice, Christopher Bryant, and Ted Briscoe. 2016.
\newblock \href {https://aclanthology.org/C16-1079} {Automatic extraction of
  learner errors in {ESL} sentences using linguistically enhanced alignments}.
\newblock In \emph{Proceedings of {COLING} 2016, the 26th International
  Conference on Computational Linguistics: Technical Papers}, pages 825--835,
  Osaka, Japan. The COLING 2016 Organizing Committee.

\bibitem[{Feng et~al.(2023{\natexlab{a}})Feng, Qiang, Li, Yuan, and
  Zhu}]{feng2023sentence}
Yutao Feng, Jipeng Qiang, Yun Li, Yunhao Yuan, and Yi~Zhu. 2023{\natexlab{a}}.
\newblock Sentence simplification via large language models.
\newblock \emph{arXiv preprint arXiv:2302.11957}.

\bibitem[{Feng et~al.(2023{\natexlab{b}})Feng, Qiang, Li, Yuan, and
  Zhu}]{Feng2023SentenceSV}
Yutao Feng, Jipeng Qiang, Yun Li, Yunhao Yuan, and Yi~Zhu. 2023{\natexlab{b}}.
\newblock Sentence simplification via large language models.
\newblock \emph{ArXiv}, abs/2302.11957.

\bibitem[{Gu et~al.(2019)Gu, Wang, and Zhao}]{gu2019levenshtein}
Jiatao Gu, Changhan Wang, and Junbo Zhao. 2019.
\newblock Levenshtein transformer.
\newblock \emph{Advances in Neural Information Processing Systems}, 32.

\bibitem[{Guu et~al.(2018)Guu, Hashimoto, Oren, and
  Liang}]{guu-etal-2018-generating}
Kelvin Guu, Tatsunori~B. Hashimoto, Yonatan Oren, and Percy Liang. 2018.
\newblock \href {https://doi.org/10.1162/tacl_a_00030} {Generating sentences by
  editing prototypes}.
\newblock \emph{Transactions of the Association for Computational Linguistics},
  6:437--450.

\bibitem[{Kaneko et~al.(2020)Kaneko, Mita, Kiyono, Suzuki, and
  Inui}]{kaneko-etal-2020-encoder}
Masahiro Kaneko, Masato Mita, Shun Kiyono, Jun Suzuki, and Kentaro Inui. 2020.
\newblock \href {https://doi.org/10.18653/v1/2020.acl-main.391}
  {Encoder-decoder models can benefit from pre-trained masked language models
  in grammatical error correction}.
\newblock In \emph{Proceedings of the 58th Annual Meeting of the Association
  for Computational Linguistics}, pages 4248--4254, Online. Association for
  Computational Linguistics.

\bibitem[{Kaneko and Okazaki(2023)}]{kaneko2023controlled}
Masahiro Kaneko and Naoaki Okazaki. 2023.
\newblock Controlled generation with prompt insertion for natural language
  explanations in grammatical error correction.
\newblock \emph{arXiv preprint arXiv:2309.11439}.

\bibitem[{Kaneko et~al.(2022)Kaneko, Takase, Niwa, and
  Okazaki}]{kaneko-etal-2022-interpretability}
Masahiro Kaneko, Sho Takase, Ayana Niwa, and Naoaki Okazaki. 2022.
\newblock \href {https://doi.org/10.18653/v1/2022.acl-long.496}
  {Interpretability for language learners using example-based grammatical error
  correction}.
\newblock In \emph{Proceedings of the 60th Annual Meeting of the Association
  for Computational Linguistics (Volume 1: Long Papers)}, pages 7176--7187,
  Dublin, Ireland. Association for Computational Linguistics.

\bibitem[{Kaplan et~al.(2020)Kaplan, McCandlish, Henighan, Brown, Chess, Child,
  Gray, Radford, Wu, and Amodei}]{kaplan2020scaling}
Jared Kaplan, Sam McCandlish, Tom Henighan, Tom~B Brown, Benjamin Chess, Rewon
  Child, Scott Gray, Alec Radford, Jeffrey Wu, and Dario Amodei. 2020.
\newblock Scaling laws for neural language models.
\newblock \emph{arXiv preprint arXiv:2001.08361}.

\bibitem[{Kitaev et~al.(2020)Kitaev, Kaiser, and Levskaya}]{kitaev2020reformer}
Nikita Kitaev, {\L}ukasz Kaiser, and Anselm Levskaya. 2020.
\newblock Reformer: The efficient transformer.
\newblock \emph{arXiv preprint arXiv:2001.04451}.

\bibitem[{Koide et~al.(2018)Koide, Kawano, and Kutsuna}]{koide2018neural}
Satoshi Koide, Keisuke Kawano, and Takuro Kutsuna. 2018.
\newblock Neural edit operations for biological sequences.
\newblock \emph{Advances in Neural Information Processing Systems}, 31.

\bibitem[{Kraus et~al.(2023)Kraus, Bingler, Leippold, Schimanski, Senni,
  Stammbach, Vaghefi, and Webersinke}]{kraus2023enhancing}
Mathias Kraus, Julia~Anna Bingler, Markus Leippold, Tobias Schimanski,
  Chiara~Colesanti Senni, Dominik Stammbach, Saeid~Ashraf Vaghefi, and Nicolas
  Webersinke. 2023.
\newblock Enhancing large language models with climate resources.
\newblock \emph{arXiv preprint arXiv:2304.00116}.

\bibitem[{Kumar et~al.(2020)Kumar, Ahuja, Vadapalli, and
  Talukdar}]{10.1162/tacl_a_00318}
Ashutosh Kumar, Kabir Ahuja, Raghuram Vadapalli, and Partha Talukdar. 2020.
\newblock \href {https://doi.org/10.1162/tacl_a_00318} {{Syntax-Guided
  Controlled Generation of Paraphrases}}.
\newblock \emph{Transactions of the Association for Computational Linguistics},
  8:330--345.

\bibitem[{Lai et~al.(2021)Lai, Toral, and Nissim}]{lai-etal-2021-thank}
Huiyuan Lai, Antonio Toral, and Malvina Nissim. 2021.
\newblock \href {https://doi.org/10.18653/v1/2021.acl-short.62} {Thank you
  {BART}! rewarding pre-trained models improves formality style transfer}.
\newblock In \emph{Proceedings of the 59th Annual Meeting of the Association
  for Computational Linguistics and the 11th International Joint Conference on
  Natural Language Processing (Volume 2: Short Papers)}, pages 484--494,
  Online. Association for Computational Linguistics.

\bibitem[{Li et~al.(2022)Li, Wang, Fan, Wang, Li, and
  Wang}]{li-etal-2022-learning-adapt}
Zhigen Li, Yanmeng Wang, Rizhao Fan, Ye~Wang, Jianfeng Li, and Shaojun Wang.
  2022.
\newblock \href {https://aclanthology.org/2022.emnlp-main.66} {Learning to
  adapt to low-resource paraphrase generation}.
\newblock In \emph{Proceedings of the 2022 Conference on Empirical Methods in
  Natural Language Processing}, pages 1014--1022, Abu Dhabi, United Arab
  Emirates. Association for Computational Linguistics.

\bibitem[{Lin(2004)}]{lin-2004-rouge}
Chin-Yew Lin. 2004.
\newblock \href {https://aclanthology.org/W04-1013} {{ROUGE}: A package for
  automatic evaluation of summaries}.
\newblock In \emph{Text Summarization Branches Out}, pages 74--81, Barcelona,
  Spain. Association for Computational Linguistics.

\bibitem[{Liu et~al.(2022)Liu, Wang, and Okazaki}]{liu-etal-2022-semi}
Ao~Liu, An~Wang, and Naoaki Okazaki. 2022.
\newblock \href {https://doi.org/10.18653/v1/2022.acl-long.321}
  {Semi-supervised formality style transfer with consistency training}.
\newblock In \emph{Proceedings of the 60th Annual Meeting of the Association
  for Computational Linguistics (Volume 1: Long Papers)}, pages 4689--4701,
  Dublin, Ireland. Association for Computational Linguistics.

\bibitem[{Liu et~al.(2018)Liu, Saleh, Pot, Goodrich, Sepassi, Kaiser, and
  Shazeer}]{liu2018generating}
Peter~J Liu, Mohammad Saleh, Etienne Pot, Ben Goodrich, Ryan Sepassi, Lukasz
  Kaiser, and Noam Shazeer. 2018.
\newblock Generating wikipedia by summarizing long sequences.
\newblock \emph{arXiv preprint arXiv:1801.10198}.

\bibitem[{Liu et~al.(2019)Liu, Ott, Goyal, Du, Joshi, Chen, Levy, Lewis,
  Zettlemoyer, and Stoyanov}]{liu2019roberta}
Yinhan Liu, Myle Ott, Naman Goyal, Jingfei Du, Mandar Joshi, Danqi Chen, Omer
  Levy, Mike Lewis, Luke Zettlemoyer, and Veselin Stoyanov. 2019.
\newblock Roberta: A robustly optimized bert pretraining approach.
\newblock \emph{arXiv preprint arXiv:1907.11692}.

\bibitem[{Loem et~al.(2023)Loem, Kaneko, Takase, and
  Okazaki}]{loem2023exploring}
Mengsay Loem, Masahiro Kaneko, Sho Takase, and Naoaki Okazaki. 2023.
\newblock Exploring effectiveness of gpt-3 in grammatical error correction: A
  study on performance and controllability in prompt-based methods.
\newblock \emph{arXiv preprint arXiv:2305.18156}.

\bibitem[{Loem et~al.(2022)Loem, Takase, Kaneko, and
  Okazaki}]{loem2022neighbors}
Mengsay Loem, Sho Takase, Masahiro Kaneko, and Naoaki Okazaki. 2022.
\newblock Are neighbors enough? multi-head neural n-gram can be alternative to
  self-attention.
\newblock \emph{arXiv preprint arXiv:2207.13354}.

\bibitem[{Mallinson et~al.(2020)Mallinson, Severyn, Malmi, and
  Garrido}]{mallinson-etal-2020-felix}
Jonathan Mallinson, Aliaksei Severyn, Eric Malmi, and Guillermo Garrido. 2020.
\newblock \href {https://doi.org/10.18653/v1/2020.findings-emnlp.111} {{FELIX}:
  Flexible text editing through tagging and insertion}.
\newblock In \emph{Findings of the Association for Computational Linguistics:
  EMNLP 2020}, pages 1244--1255, Online. Association for Computational
  Linguistics.

\bibitem[{Malmi et~al.(2020)Malmi, Severyn, and
  Rothe}]{malmi-etal-2020-unsupervised}
Eric Malmi, Aliaksei Severyn, and Sascha Rothe. 2020.
\newblock \href {https://doi.org/10.18653/v1/2020.emnlp-main.699} {Unsupervised
  text style transfer with padded masked language models}.
\newblock In \emph{Proceedings of the 2020 Conference on Empirical Methods in
  Natural Language Processing (EMNLP)}, pages 8671--8680, Online. Association
  for Computational Linguistics.

\bibitem[{Martin et~al.(2020)Martin, de~la Clergerie, Sagot, and
  Bordes}]{martin-etal-2020-controllable}
Louis Martin, {\'E}ric de~la Clergerie, Beno{\^\i}t Sagot, and Antoine Bordes.
  2020.
\newblock \href {https://aclanthology.org/2020.lrec-1.577} {Controllable
  sentence simplification}.
\newblock In \emph{Proceedings of the Twelfth Language Resources and Evaluation
  Conference}, pages 4689--4698, Marseille, France. European Language Resources
  Association.

\bibitem[{Martin et~al.(2022)Martin, Fan, de~la Clergerie, Bordes, and
  Sagot}]{martin-etal-2022-muss}
Louis Martin, Angela Fan, {\'E}ric de~la Clergerie, Antoine Bordes, and
  Beno{\^\i}t Sagot. 2022.
\newblock \href {https://aclanthology.org/2022.lrec-1.176} {{MUSS}:
  Multilingual unsupervised sentence simplification by mining paraphrases}.
\newblock In \emph{Proceedings of the Thirteenth Language Resources and
  Evaluation Conference}, pages 1651--1664, Marseille, France. European
  Language Resources Association.

\bibitem[{Meng et~al.(2021)Meng, Ao, He, Sun, Han, Wu, Fan, and
  Li}]{meng-etal-2021-conrpg}
Yuxian Meng, Xiang Ao, Qing He, Xiaofei Sun, Qinghong Han, Fei Wu, Chun Fan,
  and Jiwei Li. 2021.
\newblock \href {https://doi.org/10.18653/v1/2021.emnlp-main.199} {{C}on{RPG}:
  Paraphrase generation using contexts as regularizer}.
\newblock In \emph{Proceedings of the 2021 Conference on Empirical Methods in
  Natural Language Processing}, pages 2551--2562, Online and Punta Cana,
  Dominican Republic. Association for Computational Linguistics.

\bibitem[{Ng et~al.(2014)Ng, Wu, Briscoe, Hadiwinoto, Susanto, and
  Bryant}]{ng-etal-2014-conll}
Hwee~Tou Ng, Siew~Mei Wu, Ted Briscoe, Christian Hadiwinoto, Raymond~Hendy
  Susanto, and Christopher Bryant. 2014.
\newblock \href {https://doi.org/10.3115/v1/W14-1701} {The {C}o{NLL}-2014
  shared task on grammatical error correction}.
\newblock In \emph{Proceedings of the Eighteenth Conference on Computational
  Natural Language Learning: Shared Task}, pages 1--14, Baltimore, Maryland.
  Association for Computational Linguistics.

\bibitem[{Ng et~al.(2013)Ng, Wu, Wu, Hadiwinoto, and
  Tetreault}]{ng-etal-2013-conll}
Hwee~Tou Ng, Siew~Mei Wu, Yuanbin Wu, Christian Hadiwinoto, and Joel Tetreault.
  2013.
\newblock \href {https://aclanthology.org/W13-3601} {The {C}o{NLL}-2013 shared
  task on grammatical error correction}.
\newblock In \emph{Proceedings of the Seventeenth Conference on Computational
  Natural Language Learning: Shared Task}, pages 1--12, Sofia, Bulgaria.
  Association for Computational Linguistics.

\bibitem[{Omelianchuk et~al.(2020)Omelianchuk, Atrasevych, Chernodub, and
  Skurzhanskyi}]{omelianchuk-etal-2020-gector}
Kostiantyn Omelianchuk, Vitaliy Atrasevych, Artem Chernodub, and Oleksandr
  Skurzhanskyi. 2020.
\newblock \href {https://doi.org/10.18653/v1/2020.bea-1.16} {{GECT}o{R} {--}
  grammatical error correction: Tag, not rewrite}.
\newblock In \emph{Proceedings of the Fifteenth Workshop on Innovative Use of
  NLP for Building Educational Applications}, pages 163--170, Seattle, WA, USA
  → Online. Association for Computational Linguistics.

\bibitem[{OpenAI(2023)}]{chatgpt}
OpenAI. 2023.
\newblock \href {https://openai.com/blog/chatgpt} {Introducing {ChatGPT}}.

\bibitem[{Ouyang et~al.(2022)Ouyang, Wu, Jiang, Almeida, Wainwright, Mishkin,
  Zhang, Agarwal, Slama, Ray et~al.}]{ouyang2022training}
Long Ouyang, Jeffrey Wu, Xu~Jiang, Diogo Almeida, Carroll Wainwright, Pamela
  Mishkin, Chong Zhang, Sandhini Agarwal, Katarina Slama, Alex Ray, et~al.
  2022.
\newblock Training language models to follow instructions with human feedback.
\newblock \emph{Advances in Neural Information Processing Systems},
  35:27730--27744.

\bibitem[{Papineni et~al.(2002)Papineni, Roukos, Ward, and
  Zhu}]{papineni2002bleu}
Kishore Papineni, Salim Roukos, Todd Ward, and Wei-Jing Zhu. 2002.
\newblock Bleu: a method for automatic evaluation of machine translation.
\newblock In \emph{Proceedings of the 40th annual meeting of the Association
  for Computational Linguistics}, pages 311--318.

\bibitem[{Peng et~al.(2021)Peng, Pappas, Yogatama, Schwartz, Smith, and
  Kong}]{peng2021random}
Hao Peng, Nikolaos Pappas, Dani Yogatama, Roy Schwartz, Noah~A Smith, and
  Lingpeng Kong. 2021.
\newblock Random feature attention.
\newblock \emph{arXiv preprint arXiv:2103.02143}.

\bibitem[{Qiu et~al.(2020)Qiu, Ma, Levy, Yih, Wang, and
  Tang}]{qiu-etal-2020-blockwise}
Jiezhong Qiu, Hao Ma, Omer Levy, Wen-tau Yih, Sinong Wang, and Jie Tang. 2020.
\newblock \href {https://doi.org/10.18653/v1/2020.findings-emnlp.232}
  {Blockwise self-attention for long document understanding}.
\newblock In \emph{Findings of the Association for Computational Linguistics:
  EMNLP 2020}, pages 2555--2565, Online. Association for Computational
  Linguistics.

\bibitem[{Qorib et~al.(2022)Qorib, Na, and Ng}]{qorib-etal-2022-frustratingly}
Muhammad Qorib, Seung-Hoon Na, and Hwee~Tou Ng. 2022.
\newblock \href {https://doi.org/10.18653/v1/2022.naacl-main.143}
  {Frustratingly easy system combination for grammatical error correction}.
\newblock In \emph{Proceedings of the 2022 Conference of the North American
  Chapter of the Association for Computational Linguistics: Human Language
  Technologies}, pages 1964--1974, Seattle, United States. Association for
  Computational Linguistics.

\bibitem[{Radford et~al.(2019)Radford, Wu, Child, Luan, Amodei, Sutskever
  et~al.}]{radford2019language}
Alec Radford, Jeffrey Wu, Rewon Child, David Luan, Dario Amodei, Ilya
  Sutskever, et~al. 2019.
\newblock Language models are unsupervised multitask learners.
\newblock \emph{OpenAI blog}, 1(8):9.

\bibitem[{Rao and Tetreault(2018)}]{rao-tetreault-2018-dear}
Sudha Rao and Joel Tetreault. 2018.
\newblock \href {https://doi.org/10.18653/v1/N18-1012} {Dear sir or madam, may
  {I} introduce the {GYAFC} dataset: Corpus, benchmarks and metrics for
  formality style transfer}.
\newblock In \emph{Proceedings of the 2018 Conference of the North {A}merican
  Chapter of the Association for Computational Linguistics: Human Language
  Technologies, Volume 1 (Long Papers)}, pages 129--140, New Orleans,
  Louisiana. Association for Computational Linguistics.

\bibitem[{Rastogi et~al.(2016)Rastogi, Cotterell, and
  Eisner}]{rastogi-etal-2016-weighting}
Pushpendre Rastogi, Ryan Cotterell, and Jason Eisner. 2016.
\newblock \href {https://doi.org/10.18653/v1/N16-1076} {Weighting finite-state
  transductions with neural context}.
\newblock In \emph{Proceedings of the 2016 Conference of the North {A}merican
  Chapter of the Association for Computational Linguistics: Human Language
  Technologies}, pages 623--633, San Diego, California. Association for
  Computational Linguistics.

\bibitem[{Reid and Neubig(2022)}]{reid-neubig-2022-learning}
Machel Reid and Graham Neubig. 2022.
\newblock \href {https://aclanthology.org/2022.findings-emnlp.280} {Learning to
  model editing processes}.
\newblock In \emph{Findings of the Association for Computational Linguistics:
  EMNLP 2022}, pages 3822--3832, Abu Dhabi, United Arab Emirates. Association
  for Computational Linguistics.

\bibitem[{Reif et~al.(2022)Reif, Ippolito, Yuan, Coenen, Callison-Burch, and
  Wei}]{reif-etal-2022-recipe}
Emily Reif, Daphne Ippolito, Ann Yuan, Andy Coenen, Chris Callison-Burch, and
  Jason Wei. 2022.
\newblock \href {https://doi.org/10.18653/v1/2022.acl-short.94} {A recipe for
  arbitrary text style transfer with large language models}.
\newblock In \emph{Proceedings of the 60th Annual Meeting of the Association
  for Computational Linguistics (Volume 2: Short Papers)}, pages 837--848,
  Dublin, Ireland. Association for Computational Linguistics.

\bibitem[{Ribeiro et~al.(2018)Ribeiro, Narayan, Cohen, and
  Carreras}]{ribeiro-etal-2018-local}
Joana Ribeiro, Shashi Narayan, Shay~B. Cohen, and Xavier Carreras. 2018.
\newblock \href {https://aclanthology.org/C18-1115} {Local string transduction
  as sequence labeling}.
\newblock In \emph{Proceedings of the 27th International Conference on
  Computational Linguistics}, pages 1360--1371, Santa Fe, New Mexico, USA.
  Association for Computational Linguistics.

\bibitem[{Rothe et~al.(2021)Rothe, Mallinson, Malmi, Krause, and
  Severyn}]{rothe-etal-2021-simple}
Sascha Rothe, Jonathan Mallinson, Eric Malmi, Sebastian Krause, and Aliaksei
  Severyn. 2021.
\newblock \href {https://doi.org/10.18653/v1/2021.acl-short.89} {A simple
  recipe for multilingual grammatical error correction}.
\newblock In \emph{Proceedings of the 59th Annual Meeting of the Association
  for Computational Linguistics and the 11th International Joint Conference on
  Natural Language Processing (Volume 2: Short Papers)}, pages 702--707,
  Online. Association for Computational Linguistics.

\bibitem[{Roy et~al.(2021)Roy, Saffar, Vaswani, and
  Grangier}]{roy-etal-2021-efficient}
Aurko Roy, Mohammad Saffar, Ashish Vaswani, and David Grangier. 2021.
\newblock \href {https://doi.org/10.1162/tacl_a_00353} {Efficient content-based
  sparse attention with routing transformers}.
\newblock \emph{Transactions of the Association for Computational Linguistics},
  9:53--68.

\bibitem[{Scao et~al.(2022)Scao, Fan, Akiki, Pavlick, Ili{\'c}, Hesslow,
  Castagn{\'e}, Luccioni, Yvon, Gall{\'e} et~al.}]{scao2022bloom}
Teven~Le Scao, Angela Fan, Christopher Akiki, Ellie Pavlick, Suzana Ili{\'c},
  Daniel Hesslow, Roman Castagn{\'e}, Alexandra~Sasha Luccioni, Fran{\c{c}}ois
  Yvon, Matthias Gall{\'e}, et~al. 2022.
\newblock Bloom: A 176b-parameter open-access multilingual language model.
\newblock \emph{arXiv preprint arXiv:2211.05100}.

\bibitem[{Schnober et~al.(2016)Schnober, Eger, Do~Dinh, and
  Gurevych}]{schnober-etal-2016-still}
Carsten Schnober, Steffen Eger, Erik-L{\^a}n Do~Dinh, and Iryna Gurevych. 2016.
\newblock \href {https://aclanthology.org/C16-1160} {Still not there? comparing
  traditional sequence-to-sequence models to encoder-decoder neural networks on
  monotone string translation tasks}.
\newblock In \emph{Proceedings of {COLING} 2016, the 26th International
  Conference on Computational Linguistics: Technical Papers}, pages 1703--1714,
  Osaka, Japan. The COLING 2016 Organizing Committee.

\bibitem[{Sukhbaatar et~al.(2019)Sukhbaatar, Grave, Lample, Jegou, and
  Joulin}]{sukhbaatar2019augmenting}
Sainbayar Sukhbaatar, Edouard Grave, Guillaume Lample, Herve Jegou, and Armand
  Joulin. 2019.
\newblock Augmenting self-attention with persistent memory.
\newblock \emph{arXiv preprint arXiv:1907.01470}.

\bibitem[{Sun et~al.(2022)Sun, Ge, Ma, Li, Wei, and Wang}]{sun2022unified}
Xin Sun, Tao Ge, Shuming Ma, Jingjing Li, Furu Wei, and Houfeng Wang. 2022.
\newblock A unified strategy for multilingual grammatical error correction with
  pre-trained cross-lingual language model.
\newblock \emph{arXiv preprint arXiv:2201.10707}.

\bibitem[{Taori et~al.(2023)Taori, Gulrajani, Zhang, Dubois, Li, Guestrin,
  Liang, and Hashimoto}]{alpaca}
Rohan Taori, Ishaan Gulrajani, Tianyi Zhang, Yann Dubois, Xuechen Li, Carlos
  Guestrin, Percy Liang, and Tatsunori~B. Hashimoto. 2023.
\newblock Stanford alpaca: An instruction-following llama model.
\newblock \url{https://github.com/tatsu-lab/stanford_alpaca}.

\bibitem[{Tay et~al.(2020)Tay, Bahri, Yang, Metzler, and Juan}]{tay2020sparse}
Yi~Tay, Dara Bahri, Liu Yang, Donald Metzler, and Da-Cheng Juan. 2020.
\newblock Sparse sinkhorn attention.
\newblock In \emph{International Conference on Machine Learning}, pages
  9438--9447. PMLR.

\bibitem[{Tay et~al.(2022)Tay, Dehghani, Bahri, and Metzler}]{tay2022efficient}
Yi~Tay, Mostafa Dehghani, Dara Bahri, and Donald Metzler. 2022.
\newblock Efficient transformers: A survey.
\newblock \emph{ACM Computing Surveys}, 55(6):1--28.

\bibitem[{Team(2023)}]{MosaicML2023Introducing}
MosaicML~NLP Team. 2023.
\newblock \href {www.mosaicml.com/blog/mpt-7b} {Introducing mpt-7b: A new
  standard for open-source, ly usable llms}.

\bibitem[{Touvron et~al.(2023)Touvron, Lavril, Izacard, Martinet, Lachaux,
  Lacroix, Rozi{\`e}re, Goyal, Hambro, Azhar, Rodriguez, Joulin, Grave, and
  Lample}]{touvron2023llama}
Hugo Touvron, Thibaut Lavril, Gautier Izacard, Xavier Martinet, Marie-Anne
  Lachaux, Timoth{\'e}e Lacroix, Baptiste Rozi{\`e}re, Naman Goyal, Eric
  Hambro, Faisal Azhar, Aurelien Rodriguez, Armand Joulin, Edouard Grave, and
  Guillaume Lample. 2023.
\newblock Llama: Open and efficient foundation language models.
\newblock \emph{arXiv preprint arXiv:2302.13971}.

\bibitem[{Vyas et~al.(2020)Vyas, Katharopoulos, and Fleuret}]{vyas2020fast}
Apoorv Vyas, Angelos Katharopoulos, and Fran{\c{c}}ois Fleuret. 2020.
\newblock Fast transformers with clustered attention.
\newblock \emph{Advances in Neural Information Processing Systems},
  33:21665--21674.

\bibitem[{Wahle et~al.(2022)Wahle, Ruas, Kirstein, and
  Gipp}]{wahle-etal-2022-large}
Jan~Philip Wahle, Terry Ruas, Frederic Kirstein, and Bela Gipp. 2022.
\newblock \href {https://aclanthology.org/2022.emnlp-main.62} {How large
  language models are transforming machine-paraphrase plagiarism}.
\newblock In \emph{Proceedings of the 2022 Conference on Empirical Methods in
  Natural Language Processing}, pages 952--963, Abu Dhabi, United Arab
  Emirates. Association for Computational Linguistics.

\bibitem[{Wang et~al.(2021)Wang, Zhou, Gan, Chen, Fang, Sun, Cheng, and
  Liu}]{wang-etal-2021-cluster}
Shuohang Wang, Luowei Zhou, Zhe Gan, Yen-Chun Chen, Yuwei Fang, Siqi Sun,
  Yu~Cheng, and Jingjing Liu. 2021.
\newblock \href {https://doi.org/10.18653/v1/2021.findings-acl.346}
  {Cluster-former: Clustering-based sparse transformer for question answering}.
\newblock In \emph{Findings of the Association for Computational Linguistics:
  ACL-IJCNLP 2021}, pages 3958--3968, Online. Association for Computational
  Linguistics.

\bibitem[{Wang et~al.(2020)Wang, Li, Khabsa, Fang, and Ma}]{wang2020linformer}
Sinong Wang, Belinda~Z Li, Madian Khabsa, Han Fang, and Hao Ma. 2020.
\newblock Linformer: Self-attention with linear complexity.
\newblock \emph{arXiv preprint arXiv:2006.04768}.

\bibitem[{Wang et~al.(2022)Wang, Kordi, Mishra, Liu, Smith, Khashabi, and
  Hajishirzi}]{wang2022self}
Yizhong Wang, Yeganeh Kordi, Swaroop Mishra, Alisa Liu, Noah~A Smith, Daniel
  Khashabi, and Hannaneh Hajishirzi. 2022.
\newblock Self-instruct: Aligning language model with self generated
  instructions.
\newblock \emph{arXiv preprint arXiv:2212.10560}.

\bibitem[{Wei et~al.(2021)Wei, Bosma, Zhao, Guu, Yu, Lester, Du, Dai, and
  Le}]{wei2021finetuned}
Jason Wei, Maarten Bosma, Vincent~Y Zhao, Kelvin Guu, Adams~Wei Yu, Brian
  Lester, Nan Du, Andrew~M Dai, and Quoc~V Le. 2021.
\newblock Finetuned language models are zero-shot learners.
\newblock \emph{arXiv preprint arXiv:2109.01652}.

\bibitem[{Wei et~al.(2022)Wei, Tay, Bommasani, Raffel, Zoph, Borgeaud,
  Yogatama, Bosma, Zhou, Metzler et~al.}]{wei2022emergent}
Jason Wei, Yi~Tay, Rishi Bommasani, Colin Raffel, Barret Zoph, Sebastian
  Borgeaud, Dani Yogatama, Maarten Bosma, Denny Zhou, Donald Metzler, et~al.
  2022.
\newblock Emergent abilities of large language models.
\newblock \emph{arXiv preprint arXiv:2206.07682}.

\bibitem[{Witteveen and Andrews(2019)}]{witteveen-andrews-2019-paraphrasing}
Sam Witteveen and Martin Andrews. 2019.
\newblock \href {https://doi.org/10.18653/v1/D19-5623} {Paraphrasing with large
  language models}.
\newblock In \emph{Proceedings of the 3rd Workshop on Neural Generation and
  Translation}, pages 215--220, Hong Kong. Association for Computational
  Linguistics.

\bibitem[{Wu et~al.(2023{\natexlab{a}})Wu, Wang, Wan, Jiao, and
  Lyu}]{wu2023chatgpt}
Haoran Wu, Wenxuan Wang, Yuxuan Wan, Wenxiang Jiao, and Michael Lyu.
  2023{\natexlab{a}}.
\newblock Chatgpt or grammarly? evaluating chatgpt on grammatical error
  correction benchmark.
\newblock \emph{arXiv preprint arXiv:2303.13648}.

\bibitem[{Wu et~al.(2023{\natexlab{b}})Wu, Waheed, Zhang, Abdul-Mageed, and
  Aji}]{wu2023lamini}
Minghao Wu, Abdul Waheed, Chiyu Zhang, Muhammad Abdul-Mageed, and Alham~Fikri
  Aji. 2023{\natexlab{b}}.
\newblock Lamini-lm: A diverse herd of distilled models from large-scale
  instructions.
\newblock \emph{arXiv preprint arXiv:2304.14402}.

\bibitem[{Xu et~al.(2016)Xu, Napoles, Pavlick, Chen, and
  Callison-Burch}]{xu-etal-2016-optimizing}
Wei Xu, Courtney Napoles, Ellie Pavlick, Quanze Chen, and Chris Callison-Burch.
  2016.
\newblock \href {https://doi.org/10.1162/tacl_a_00107} {Optimizing statistical
  machine translation for text simplification}.
\newblock \emph{Transactions of the Association for Computational Linguistics},
  4:401--415.

\bibitem[{Xu and Carpuat(2021)}]{xu2021editor}
Weijia Xu and Marine Carpuat. 2021.
\newblock Editor: An edit-based transformer with repositioning for neural
  machine translation with soft lexical constraints.
\newblock \emph{Transactions of the Association for Computational Linguistics},
  9:311--328.

\bibitem[{Yamashita et~al.(2020)Yamashita, Katsumata, Kaneko, Imankulova, and
  Komachi}]{yamashita-etal-2020-cross}
Ikumi Yamashita, Satoru Katsumata, Masahiro Kaneko, Aizhan Imankulova, and
  Mamoru Komachi. 2020.
\newblock \href {https://doi.org/10.18653/v1/2020.coling-main.415}
  {Cross-lingual transfer learning for grammatical error correction}.
\newblock In \emph{Proceedings of the 28th International Conference on
  Computational Linguistics}, pages 4704--4715, Barcelona, Spain (Online).
  International Committee on Computational Linguistics.

\bibitem[{Zhang et~al.(2022)Zhang, Roller, Goyal, Artetxe, Chen, Chen, Dewan,
  Diab, Li, Lin et~al.}]{zhang2022opt}
Susan Zhang, Stephen Roller, Naman Goyal, Mikel Artetxe, Moya Chen, Shuohui
  Chen, Christopher Dewan, Mona Diab, Xian Li, Xi~Victoria Lin, et~al. 2022.
\newblock Opt: Open pre-trained transformer language models.
\newblock \emph{arXiv preprint arXiv:2205.01068}.

\bibitem[{Zhang et~al.(2019)Zhang, Kishore, Wu, Weinberger, and
  Artzi}]{zhang2019bertscore}
Tianyi Zhang, Varsha Kishore, Felix Wu, Kilian~Q Weinberger, and Yoav Artzi.
  2019.
\newblock Bertscore: Evaluating text generation with bert.
\newblock \emph{arXiv preprint arXiv:1904.09675}.

\bibitem[{Zhang and Lapata(2017)}]{zhang-lapata-2017-sentence}
Xingxing Zhang and Mirella Lapata. 2017.
\newblock \href {https://doi.org/10.18653/v1/D17-1062} {Sentence simplification
  with deep reinforcement learning}.
\newblock In \emph{Proceedings of the 2017 Conference on Empirical Methods in
  Natural Language Processing}, pages 584--594, Copenhagen, Denmark.
  Association for Computational Linguistics.

\bibitem[{Zhao et~al.(2023)Zhao, Zhou, Li, Tang, Wang, Hou, Min, Zhang, Zhang,
  Dong et~al.}]{zhao2023survey}
Wayne~Xin Zhao, Kun Zhou, Junyi Li, Tianyi Tang, Xiaolei Wang, Yupeng Hou,
  Yingqian Min, Beichen Zhang, Junjie Zhang, Zican Dong, et~al. 2023.
\newblock A survey of large language models.
\newblock \emph{arXiv preprint arXiv:2303.18223}.

\bibitem[{Zhu et~al.(2010)Zhu, Bernhard, and
  Gurevych}]{zhu-etal-2010-monolingual}
Zhemin Zhu, Delphine Bernhard, and Iryna Gurevych. 2010.
\newblock \href {https://aclanthology.org/C10-1152} {A monolingual tree-based
  translation model for sentence simplification}.
\newblock In \emph{Proceedings of the 23rd International Conference on
  Computational Linguistics (Coling 2010)}, pages 1353--1361, Beijing, China.
  Coling 2010 Organizing Committee.

\end{thebibliography}
\bibliographystyle{acl_natbib}

\end{document}